\documentclass[letterpaper, 10 pt, conference]{ieeeconf}  % Comment this line out if you need a4paper

\IEEEoverridecommandlockouts                              % This command is only needed if 
\usepackage[table,xcdraw]{xcolor}

\usepackage{graphicx}

\usepackage{amsmath} % assumes amsmath package installed
\usepackage{amssymb}  % assumes amsmath package installed
\usepackage{tikz}
\usepackage{array}
\newcolumntype{C}[1]{>{\centering\arraybackslash}p{#1}}
\usepackage{comment}
\usepackage{color}
\usepackage{booktabs}
\usepackage{lipsum}  
\usepackage{pifont}% http://ctan.org/pkg/pifont
\usepackage{bbm}
\usepackage{dblfloatfix}

\usepackage{caption}
\usepackage{subcaption}
\usepackage{svg}

\usepackage{algpseudocode}
\usepackage{algorithm}
\usepackage{amsmath}

\usepackage[colorlinks]{hyperref}
\usepackage[capitalize]{cleveref}
\title{\LARGE \bf
Optimizing Algorithms From Pairwise User Preferences
}

\author{Leonid Keselman$^{1}$, Katherine Shih$^{1}$, Martial Hebert$^{1}$, Aaron Steinfeld$^{1}$% <-this % stops a space
\thanks{*This work was funded by the National Institute on Disability, Independent Living, and Rehabilitation Research (90DPGE0003) and the Office of Naval Research (N00014-18-1-2503).}% <-this % stops a space
\thanks{$^{1}$Authors are with the Robotics Institute, School of Computer Science, Carnegie Mellon University, Pittsburgh PA, 15232 USA
        {\tt\small \{lkeselma,kshih,hebert,astein\}@cs.cmu.edu}}%%
}

\begin{document}

\maketitle
\thispagestyle{empty}
\pagestyle{empty}

%%%%%%%%%%%%%%%%%%%%%%%%%%%%%%%%%%%%%%%%%%%%%%%%%%%%%%%%%%%%%%%%%%%%%%%%%%%%%%%%
\begin{abstract}
Typical black-box optimization approaches in robotics focus on learning from metric scores. However, that is not always possible, as not all developers have ground truth available. Learning appropriate robot behavior in human-centric contexts often requires querying users, who typically cannot provide precise metric scores. Existing approaches leverage human feedback in an attempt to model an implicit reward function; however, this reward may be difficult or impossible to effectively capture. In this work, we introduce SortCMA to optimize algorithm parameter configurations in high dimensions based on pairwise user preferences. SortCMA efficiently and robustly leverages user input to find parameter sets without directly modeling a reward. We apply this method to tuning a commercial depth sensor without ground truth, and to robot social navigation, which involves highly complex preferences over robot behavior. We show that our method succeeds in optimizing for the user's goals and perform a user study to evaluate social navigation results.
\end{abstract}

%%%%%%%%%%%%%%%%%%%%%%%%%%%%%%%%%%%%%%%%%%%%%%%%%%%%%%%%%%%%%%%%%%%%%%%%%%%%%%%%
\section{Introduction}\label{sec:intro}
Many robotics applications require optimization over high-dimensional parameter spaces with uncertain or nonexistent ground truth. Frequently, no easily quantifiable reward signal exists. Examples of these applications can range from hardware tuning with imperfect measurements to human interaction with primarily qualitative assessments; efficient adaptation of robotics algorithms may also be useful in assistive robotics settings where users require customization~\cite{argallCustom21}. 

In these situations, we can rely on users to provide a signal of preferred behavior. Demonstrations of desired actions are often used, as in inverse reinforcement learning \cite{Abbeel2004}. However, this may require large amounts of data and/or high task loads for users asked to perform demonstrations. Other techniques actively query users for feedback to model their reward functions~\cite{Biyik2022, Fitzgerald2022}, but reliable functions describing complex human preferences can be difficult to estimate. %, using varying modalities of user input to choose new samples and update their reward. 
% Some methods instead elicit direct feedback via other user inputs. 
%However, techniques that rely upon learning a set reward function cannot necessarily express the full range of desired behavior. Human preferences may be highly nonlinear or change over time and environment. While methods exist to address these  \cite{Basu2019}, this increases computation. 

In this paper, we present an algorithm, SortCMA, that takes in user feedback to optimize parameter sets in high dimensions when no gradient exists.  SortCMA can optimize effectively when feedback is imperfect and the desired behavior is complex. It converges to acceptable solutions quickly and can be applied across many domains. To demonstrate the general usefulness of the technique, we show its application in highly disparate areas. First, we tune a highly parameterized commercial depth sensor, then show that we can achieve complex, distinctive behaviors when tuning a social navigation algorithm.

\begin{figure}[tbh]
    \centering
    \includegraphics[width=0.9\linewidth]{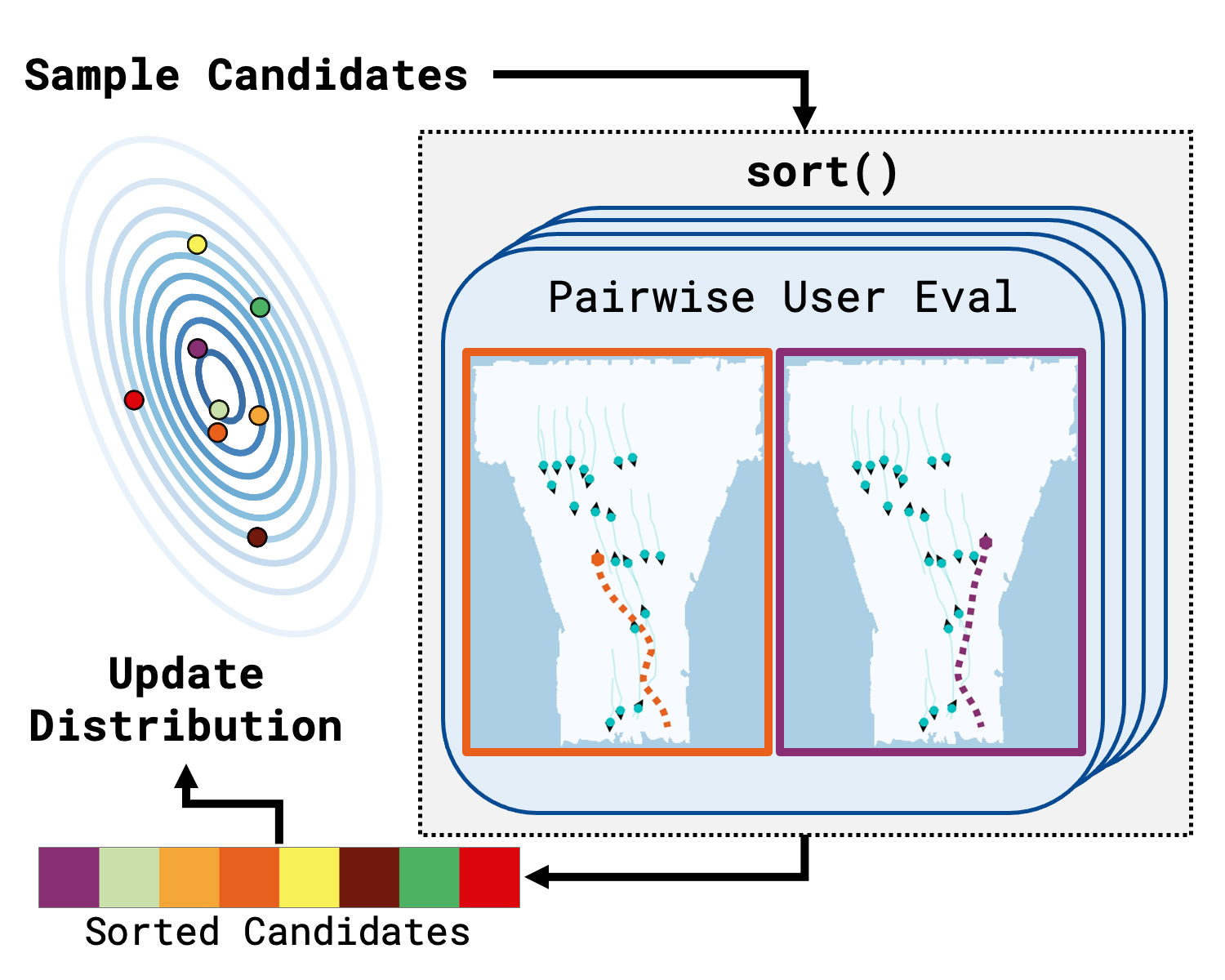}
    \caption{Our method models parameter distributions with a multivariate Gaussian and updates that model via a sorted list of preferences. The preferences are elicited from a user via A/B comparison of results. See \cref{sec:method} for details.}
    \label{fig:figure1}
\end{figure}

 \section{Related Work}\label{sec:related}
Current research on learning from user preferences explores a broad array of interaction types ranging from demonstrations~\cite{Hussein2017} to binary yes/no classification~\cite{Celemin2019}. Recent work \cite{Koppol2021} shows that of these, pairwise preference eliciting is frequently a preferred method of interaction across contexts. 
 
Of the existing works on optimizing with pairwise rewards, many use strategies that center around Bayesian Optimization (BO)~\cite{icml_chu05,JMLR:v6:chu05a}. In BO settings, a model is fit to existing samples (typically a Gaussian Process~\cite{books/lib/RasmussenW06}), and then optimized with a secondary optimizer. Classic works have been supplemented with a recent resurgence of methods~\cite{pmlr17,mlsp21,aistat22}, including those in the robotics literature~\cite{Biyik2020}. However, BO often depends on reasonable bounds to constrain the secondary optimization objective, and can struggle in high dimensional parameter spaces with extremely few queries. Our approach is focused on black-box, derivative-free, or zeroth-order optimization where we lack access to function gradients, particularly in the extreme case where the number of evaluations is very small~\cite{extreme21}. 

In our work, some experiments entirely lack metrics to model a reward function. In others, the modeled linear reward still produces behavior that is sub-optimal to preferences directly solicited from the user. Similar methods focus on other applications than algorithm tuning, and either only support a single comparison per update step~\cite{9811526}, or do not show robustness to user comparison errors~\cite{tang2023zerothorder}.

\begin{figure*}[thp]
    \smallskip
    \smallskip

    \centering
    \begin{subfigure}[t]{0.24\linewidth}
    \includegraphics[width=\linewidth]{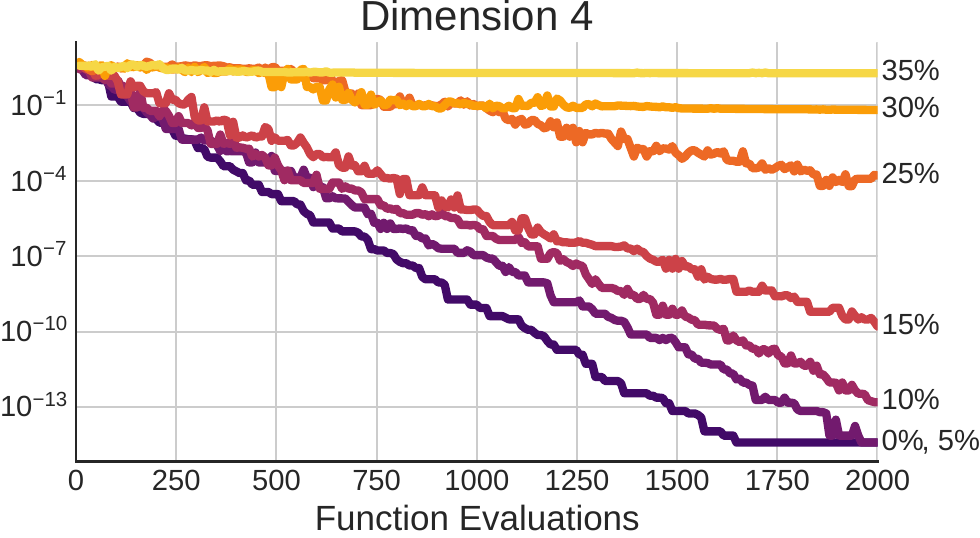} 
    \end{subfigure}
    \hfill
    \begin{subfigure}[t]{0.24\linewidth}
    \includegraphics[width=\linewidth]{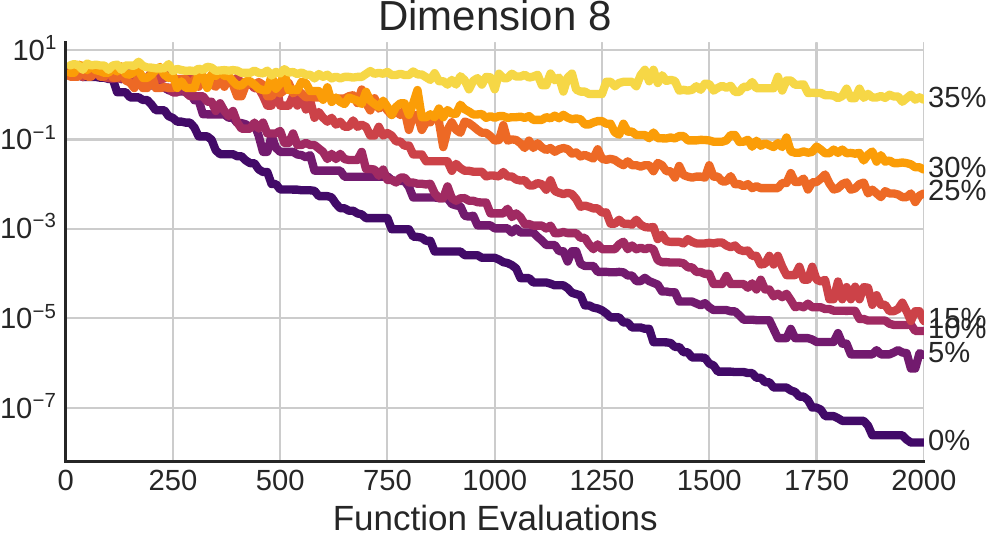} 
    \end{subfigure}
        \hfill
    \begin{subfigure}[t]{0.24\linewidth}
    \includegraphics[width=\linewidth]{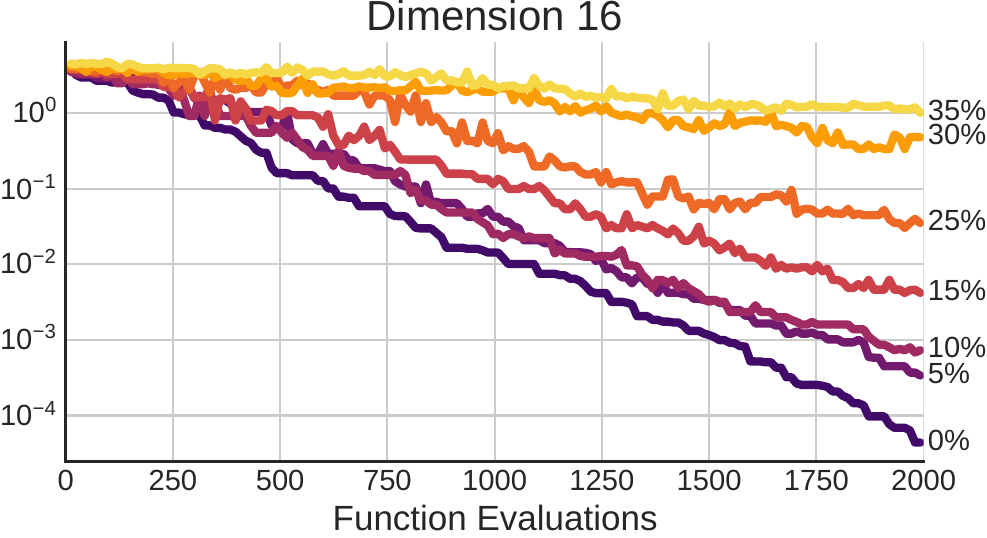} 
    \end{subfigure}
        \hfill
    \begin{subfigure}[t]{0.24\linewidth}
    \includegraphics[width=\linewidth]{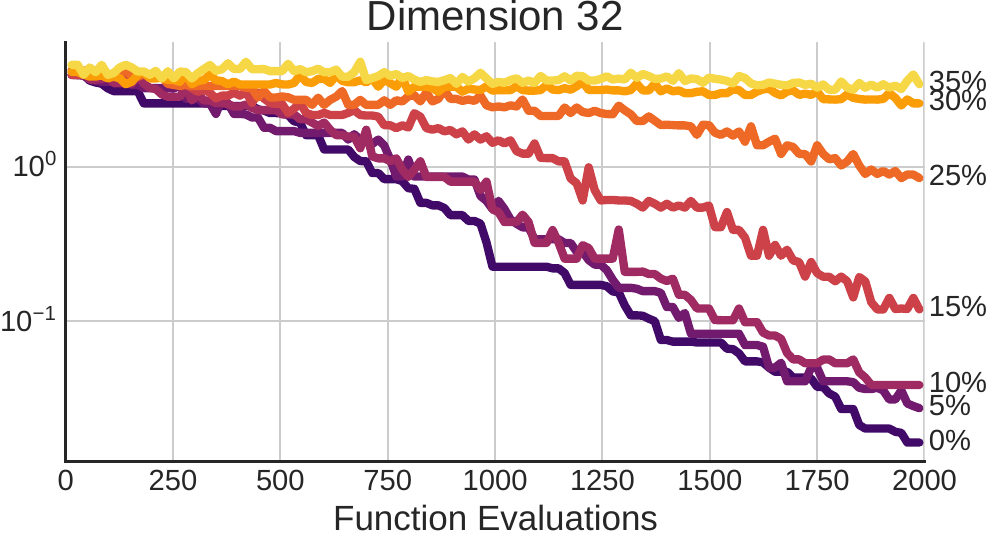} 
    \end{subfigure}
    \caption{\textbf{SortCMA with noise.} Log-loss on the Ackley test function across a range of dimensions and crossover probabilities.}
    \label{fig:noise_synth}
\end{figure*}
 \section{Method}\label{sec:method}
%Our method, which we call \textit{SortCMA}, enables efficient, robust optimization of high dimensional objective functions via user queries alone. 
Our method is designed for finding $\operatorname*{argmin}_x f(x)$, $x \in \mathbb{R}^d$, where users select either $f(x_i) \leq f(x_j)$ or $f(x_i) \geq f(x_j)$ and there is no access to $\nabla f(x)$. This represents a $d$ dimensional optimization over a series of configurations $x_{1...N}$: $f(x)$ captures user preferences, and $x$ typically represents the configuration parameters for an algorithm. We call our method SortCMA, since it builds on top of two well-studied techniques: Covariance Matrix Adaptation Evolution Strategy (CMA-ES)~\cite{hansen2016cma} and sorting methods~\cite{justsortit15}. 

 CMA-ES is a well-known, successful black-box optimization method~\cite{opt_plat}, especially effective when few function evaluations are available~\cite{extreme21}. CMA-ES has been shown to be robust and efficient enough to optimize robotic systems with humans in the loop where the optimization criteria is a continuous metric~\cite{doi:10.1126/science.aal5054}. Pairwise sorting is a classic $O(n \log n)$ algorithm with many implementations. For our results, we use Timsort. Extensions to other sorting techniques, such as radix sort, approach direct access to $f(x)$, and our method would be identical to traditional CMA-ES. 

SortCMA is a modified interface to CMA-ES that replaces direct evaluation of $f(x)$ with calling a user-querying \texttt{sort} function. The \texttt{sort}'s \texttt{key} function is a user interface that returns the user's preferred selection of $f(x_i) \leq f(x_j)$ or $f(x_i) \geq f(x_j)$. After $g$ generations, when the user is satisfied with the quality, they terminate the optimization and perform $g-1$ pairwise comparisons to select the final, optimal result.  

SortCMA and CMA-ES~\cite{hansen2016cma} are initialized with a mean $\textbf{m}^0 \in \mathbb{R}^d$, step size $\sigma^0$ and positive semi-definite covariance matrix $\textbf{C}^0 \in \mathbb{R}^{d \times d}$ (typically \textbf{I}). For each generation of the optimization, a set of $\lambda \geq 2$ offspring are sampled:

\begin{equation}
    \textbf{x}^{i + 1}_k \sim \textbf{m}^{i} + \sigma^{i}\mathcal{N}\left(\textbf{0}, \textbf{C}^i\right), \quad k = 1...\lambda .
\end{equation}

CMA-ES evaluates all $f(x^{i+1}_k), k = 1...\lambda$, and performs a sorting to assign weights $w_1 \geq w_2 \geq ... \geq w_\lambda$, where better $f(x_k)$ are assigned larger weights. Weights are defined as
\begin{equation}
    w_k = \log\left(\frac{\lambda + 1}{2}\right) - \log(k), \quad k = 1...\lambda
\end{equation}

After assigning larger weights to better $f(x_k)$, either via evaluation and sorting (CMA-ES) or direct sorting based on pairwise user evaluation (SortCMA), $\textbf{m}^{i+1}$, $\sigma^{i+1}$, and $\textbf{C}^{i+1}$  are generated using established CMA-ES update rules~\cite{hansen2016cma}. The step size ($\sigma$) is updated to control the overall exploration rate, and the covariance matrix is adapted to guide the search along favorable vectors in parameter space. 

We use \textit{pycma}~\cite{nikolaus_hansen_2023_7573532}, transform any strictly positive parameters with $f(x)=\log(x)$ for unbounded optimization, and use an initial $\sigma=0.2$ in Sec. \ref{sec:realsense} and $\sigma=0.1$ in Sec. \ref{sec:socialnav}.

\subsection{Robustness to Input Error}
When learning from user preferences, comparison errors may arise due to user uncertainty or mistakes in collecting feedback (e.g., a wrong button press). 
Theoretical developments have demonstrated that sorting methods are robust and efficient, even with noisy comparisons \cite{justsortit15,https://doi.org/10.48550/arxiv.2302.12440}. 

To evaluate the effect of potential input errors on convergence, we run SortCMA on four hard-to-optimize black-box test functions: Ackley's, Rosenbrock's, Zakharov's, and the multidimensional sphere \cite{Jamil_2013}. We study how our method behaves under the well-studied noisy comparison model with cross-over probability $p$~\cite{https://doi.org/10.48550/arxiv.2302.12440,noisy1994}. 
Each individual comparison used during sorting has a $p \in (0, \frac{1}{2})$ probability of crossover: that is, if $p = 0.25$ there is a 25{\%} probability that the disfavored option will be selected as better. Under this model of error, $p = 0.5$ gives no information. Final parameter selection is done using a noisy final sort across the best candidates from each iteration of the algorithm.

SortCMA shows convergence at exponential rates, even under noisy comparisons, in high dimension, and across a wide range of $p$ values. \Cref{fig:noise_synth} shows log-loss versus function evaluations for the Ackley function; performance was similar for the other three test functions. We found that the largest errors often came from noisy sorting in the final sort across the best of all batches. To improve overall performance, this final sort could use a modified, multi-query sorting function to achieve well-bounded behavior~\cite{https://doi.org/10.48550/arxiv.2302.12440}.

\subsection{Utility of Heuristic Option}
\begin{figure}[htbp]
\smallskip
\centering
\includegraphics[width=\linewidth]{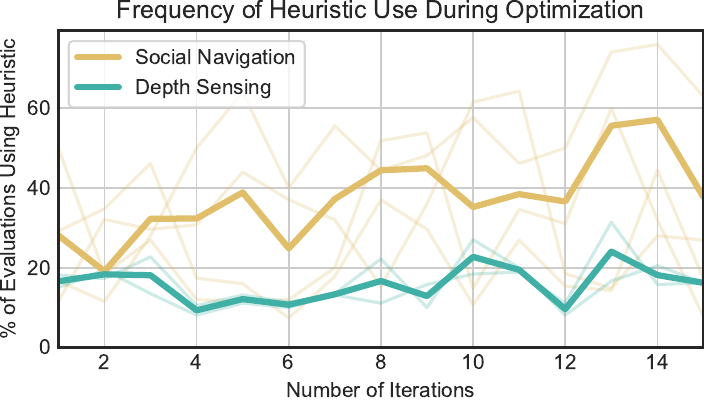}
\caption{\textbf{Reward Heuristics} are frequently used when tuning.}
\label{fig:heuristic}
\end{figure}

To assist users during optimization, we provide access to simple heuristic rewards for users to rely upon in cases where the configurations are very similar. In social navigation (\cref{sec:socialnav}), we use our metric reward, while for depth sensor tuning (\cref{sec:realsense}) we provide "more dense" and "less dense" heuristics. 
In both environments, we find that the option to allow the user to defer the pairwise choice to a heuristic is used frequently (\Cref{fig:heuristic}). In later iterations, when parameter sets are more likely to be preferred overall, this occurs nearly half of the time. Since optimization can produce example pairs that are similar, forcing a preference over all pairs can be cognitively demanding. We posit that the availability of the heuristic function reduces cognitive load for the user and is a significant factor in the usability of our method for tuning in practical applications. 

 \begin{figure*}[thpb]
  \centering
  \smallskip
  \smallskip

  \begin{subfigure}[t]{0.32\linewidth}
    \centering
    \includegraphics[width=\textwidth]{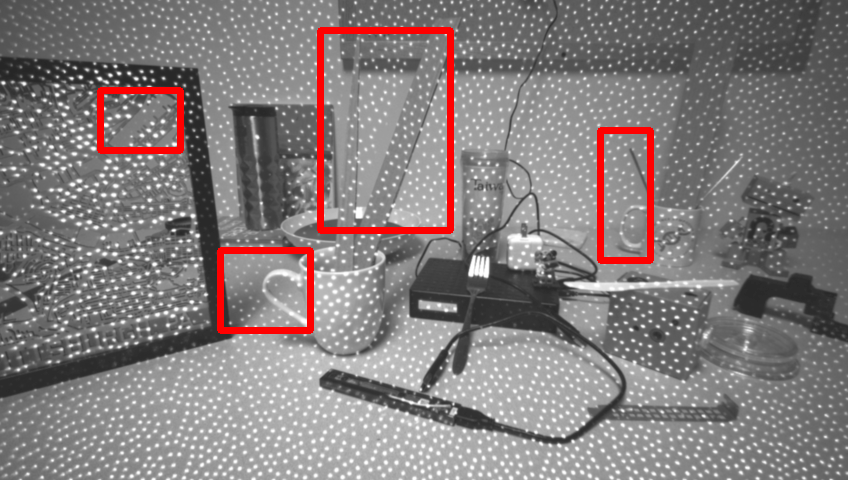}
    \caption{\textbf{Active Stereo Input}}
  \end{subfigure}
  \begin{subfigure}[t]{0.32\linewidth} 
    \centering
    \includegraphics[width=\textwidth]{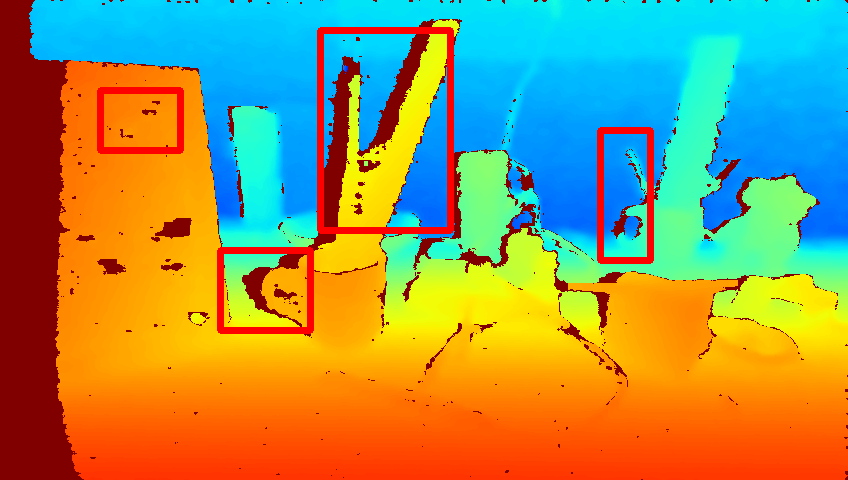}
    \caption{\textbf{Active Stereo Defaults}}
    \label{sec:active_default}
  \end{subfigure}
  \begin{subfigure}[t]{0.32\linewidth} 
    \centering
    \includegraphics[width=\textwidth]{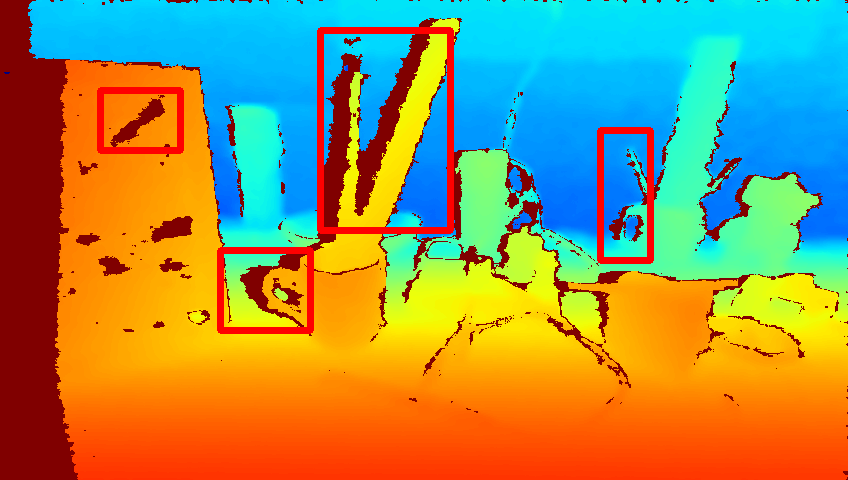}
    \caption{\textbf{Active Stereo Tuned}}
    \label{sec:active_tuned}
  \end{subfigure}
    \begin{subfigure}[t]{0.32\linewidth}
    \centering
    \includegraphics[width=\textwidth]{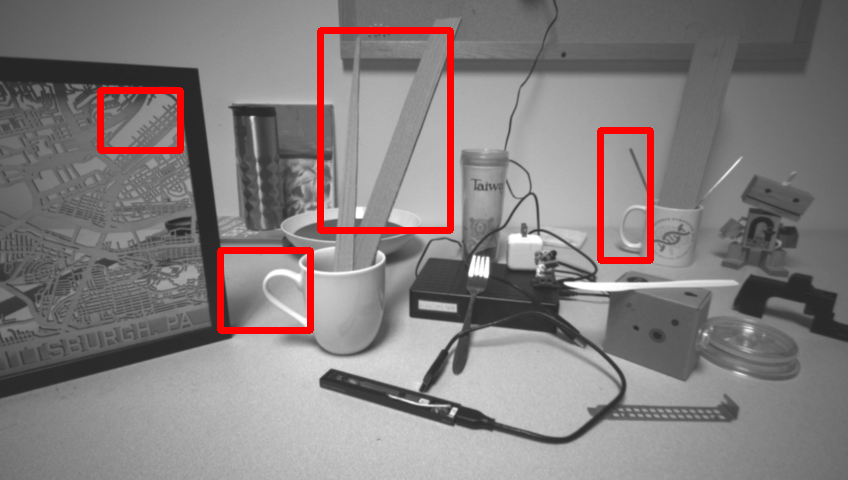}
    \caption{\textbf{Passive Stereo Input}}
  \end{subfigure}
  \begin{subfigure}[t]{0.32\linewidth} 
    \centering
    \includegraphics[width=\textwidth]{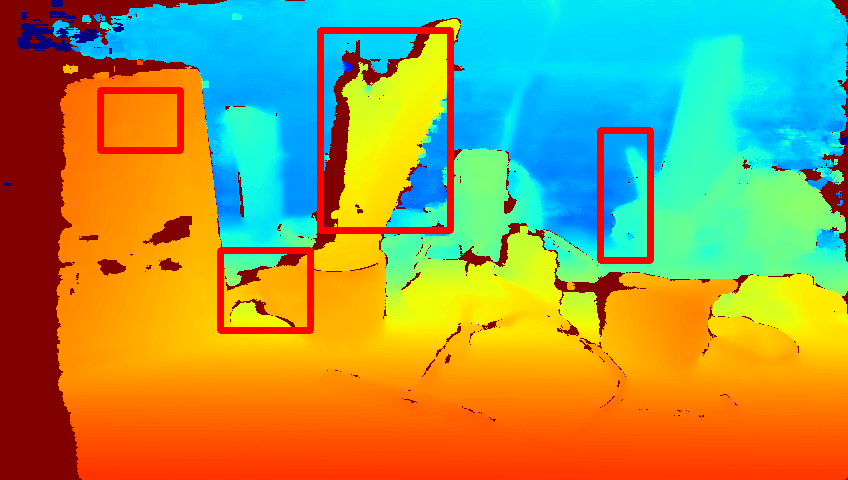}
    \caption{\textbf{Passive Stereo Defaults}}
    \label{sec:passive_default}

  \end{subfigure}
  \begin{subfigure}[t]{0.32\linewidth} 
    \centering
    \includegraphics[width=\textwidth]{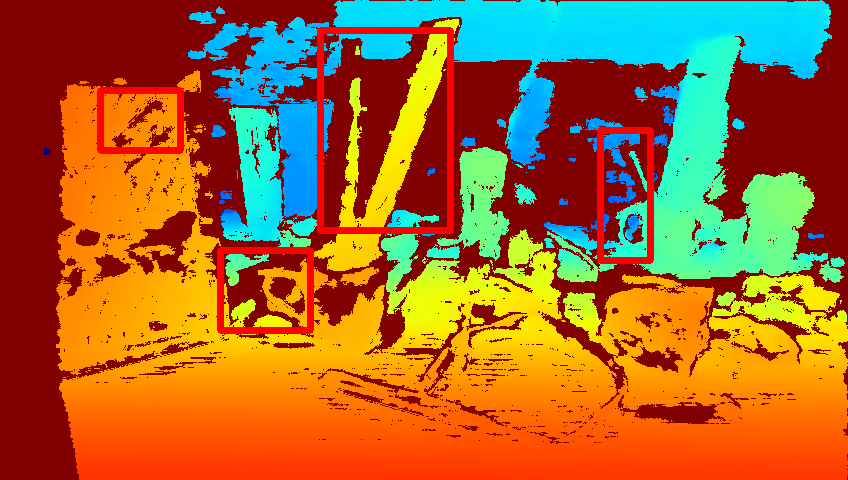}
    \caption{\textbf{Passive Stereo Tuned}}
    \label{sec:passive_tuned}

  \end{subfigure}
  \caption{\textbf{Tuning RealSense D435 Stereo Matching} As described in \cref{sec:realsense}, we use our optimizer to tune the 35 parameters that control depth generation on a commercial depth camera. The target user preference was for more precise discontinuities. The top row demonstrates dense matching with projected texture~\cite{realsense2017,prismsensor}, while the bottom row lacks texture. }
  \label{fig:stereo_out}
\end{figure*}

% \begin{algorithm}
% \caption{\strut Optimize with user preferences}\label{alg:optimizer}
% \begin{algorithmic}[1]
% \Require {$x_0$} Initial Configuration
% \Require {$f(x)$} Function to run algorithm
% \Require {$N$} Number of batches
% \Require {$M$} Configurations per batch

% \Procedure{SortCMA}{$x_0,f_(x),N$}
% \For{$i \gets 1$ to $N$}
%     \State $x_{1...M}\gets\Call{Sample}$
%     \State $j_{1...M}\gets\Call{Sort}{x_{1...M},Key=User(A,B)}$
%     \State $b_i = x_{j_M}$
%     \State $\Call{Update}{j_{1...M}}$\Comment{Update Distribution}
% \EndFor
% \State $k_{1...N}\gets\Call{Sort}{b_{1...N},Key=User(A,B)}$
% \State $x_f = b_{k_N}$ \Comment{Final Configuration}
% \EndProcedure
% \end{algorithmic}
% \end{algorithm}

\section{Tuning a Stereoscopic Depth Sensor }\label{sec:realsense}
%\subsection{Depth Sensor Setup} 
Many robots use the widely available Intel RealSense sensors~\cite{realsense2017}. The D435 is a stereoscopic depth camera that uses an ASIC-implemented depth algorithm~\cite{4359315}, along with a laser~\cite{prismsensor} for projecting additional texture. 

While it is possible to tune stereo algorithms with existing datasets~\cite{middlebury2021} or on existing camera hardware with pseudo-ground truth~\cite{keselman2023modal}, these approaches can often be lacking. In particular, real scenes will often contain pixels which are not annotated due to surface properties (specularity, transparency, etc.) or exist around edges and stereo occluded regions. Additionally, practical usage makes this a multi-objective optimization, balancing depth accuracy, fill rate and outliers (which often appear in unannotated regions). 

Instead, we show how a rich scene can be optimized with SortCMA based on, and for, user preferences. This allows optimization without designing a multi-objective loss.%The results of about 30 minutes of tuning can be seen in \Cref{fig:stereo_out}. 

\subsection{Sensor Setup} 

The RealSense depth algorithm has 35 parameters that control the properties of the algorithm implemented in hardware. The parameters control regularization of the matching algorithm and checks that ensure invalid results are discarded. By modifying these parameters, users can greatly impact the properties and characteristics of the data. %While the vendor provides several example configurations for how to configure the sensor, they do not behave well in all environments. %To showcase this, we set up a scene of rich objects where the default parameters (and all other vendor-provided parameter sets) produce clear artifacts around depth discontinuities which could be undesirable. 

We tune the depth generation for the same scene under two conditions. First, the laser emitter is enabled and dense, and high quality results are expected. In the second, the laser emitter is turned off, resulting in a difficult environment that resembles how the stereo system behaves under sunlight or at distance. These parameters are tuned directly from user preferences, without having to meticulously collect ground truth of the target scene~\cite{middlebury2021}.

 \subsection{Visual Tuning Results}\label{sec:cameras}
The results of running SortCMA on tuning a stereo depth sensor for good edge performance are shown in \cref{fig:stereo_out}. In both active and passive settings, we tune the algorithm with the default CMA-ES population size and perform optimization over 15 generations. Many approaches to pairwise optimization are unsuitable baselines in this context, as they depend on a reasonable reward function to model the final choice, but this reward function is unavailable. % ~\cite{pmlr17,Biyik2020}

We annotated four areas of interest for seeing differences between configurations: a large V shape, a hole in a woodcut, a close left mug handle and a further right mug handle. 

For stereo matching with projected texture, the initial defaults (\cref{sec:active_default}) struggle with the V shape, inpaint the woodcut hole, and cannot obtain any depth inside the left mug handle. In tuning the settings (\cref{sec:active_tuned}), we were able to  resolve the V shape, obtain better detail around the mug, and reject bad matches inside the woodcut, all without negatively affecting other parts of the scene. Optimization was quick: by the fourth generation, all samples had a clear V shape. By the sixth, the handle was sharply resolved. The last five generations mainly balanced density and outliers. 

%. Upon further inspection, clear artifacts can be seen, e.g., the \textbf{V shape} in the middle of the scene is incorrectly infilled. Smaller details are also obfuscated, such as the handle of the mug holding the V shape. 
%which resembles stereo matching in sunlight conditions, 

For passive stereo matching, the defaults are highly inpainted (\cref{sec:passive_default}) across all four selected regions of interest. The tuned  settings (\cref{sec:passive_tuned}) resolve sharp, detailed edges at the expense of dense infill in the white background areas. In this difficult case, it took five batches to obtain a clear V shape, and about ten batches until tradeoffs were tested. 

In both cases, it took about 30 minutes of tuning to get our final results, producing potentially much better configurations in a given environment. For users with particular requirements for their depth data, this could broaden and enhance the utility of commercial hardware. %Using human preferences to tune algorithms can be a viable alternative to collecting metric ground truth.

\begin{figure*}[thp]
    \smallskip
    \smallskip
    \centering
    \begin{subfigure}[t]{0.49\linewidth}
    \includegraphics[width=\linewidth]{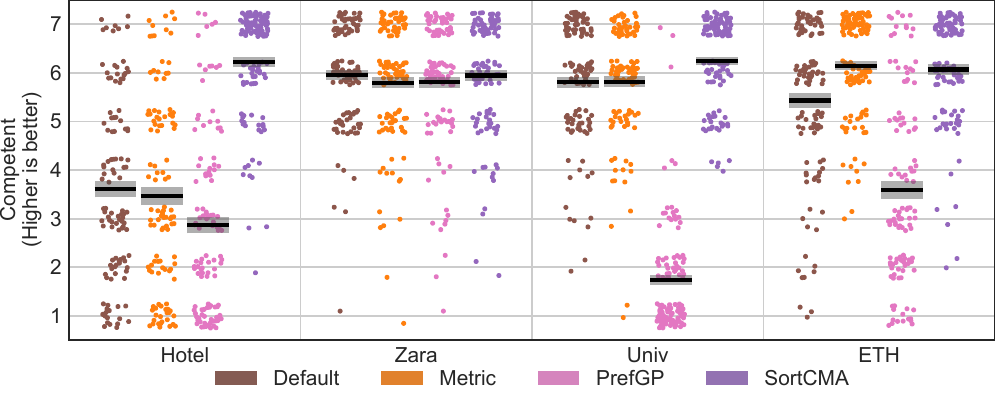} 
    \end{subfigure}
    \begin{subfigure}[t]{0.49\linewidth}
    \includegraphics[width=\linewidth]{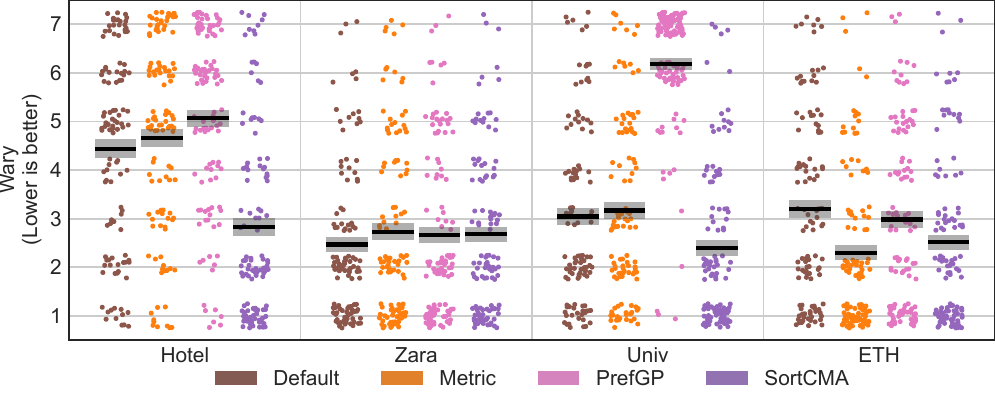} 
    \end{subfigure}
    \caption{\textbf{Evaluation of Social Navigation Policies} in our user study (\Cref{sec:userstudy}). Mean and one standard error are shown.}
    \label{fig:vote_graphs}
\end{figure*}

\section{Social Navigation}\label{sec:socialnav}
To demonstrate SortCMA in a high-complexity area with difficult to model rewards, we turn to social navigation: creating robot trajectories that conform to social norms. While robot motion planning is well-studied, generating socially appropriate robot paths is challenging. What is considered socially appropriate is mediated by culture, attitudes towards robots, and the robot’s appearance and task \cite{Mavrogiannis2023}.

We study social navigation in the context of the recent benchmark SocNavBench \cite{Biswas2022}, a simulation environment that contains pedestrian interaction scenarios sourced from the ETH and UCY datasets \cite{Lerner2007, Pellegrini2009}. A scenario in SocNavBench consists of a map, a set of real pedestrian paths, and start/goal locations for the robot. The output when running a given algorithm on a scenario is a top-down video of the robot following a trajectory, including markers of pedestrian collisions, as well as numerous performance metrics ranging from path length to distance from pedestrians. Despite this array of metrics, there is no definition of `best'. Out of the scenarios in the benchmark, we focus on a set of four, each with a different map and dense crowds.

\subsection{Navigation Tuning}\label{sec:applications}
\begin{figure*}[htb]
    \centering
    \begin{subfigure}[t]{0.9\linewidth}
    \includegraphics[width=\linewidth]{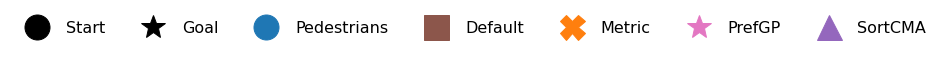} %\caption{$t$ = 4s}
    \end{subfigure}
    %\vspace{0.1cm}    
    % \begin{subfigure}[t]{0.19\linewidth}
    %\includegraphics[width=\linewidth]{figs/vote_graph_new/Hotel_pair.png} %\caption{$t$ = 3.25s}
    %\end{subfigure}
    \begin{subfigure}[t]{0.19\linewidth}
    \includegraphics[width=\linewidth]{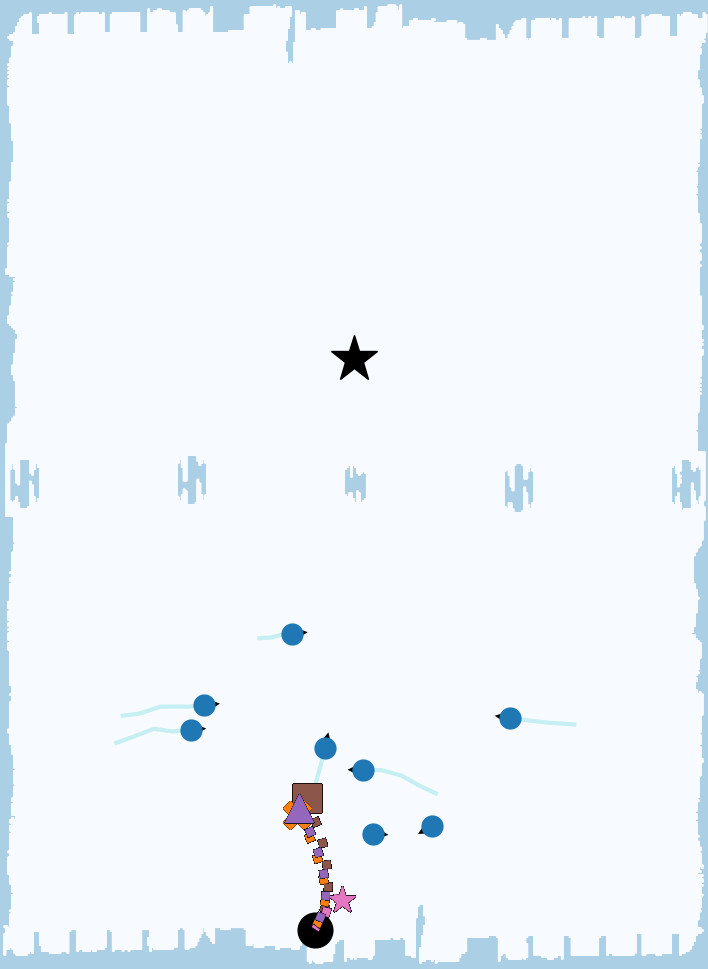} %\caption{$t$ = 3.25s}
    \end{subfigure}
    \begin{subfigure}[t]{0.19\linewidth}
    \includegraphics[width=\linewidth]{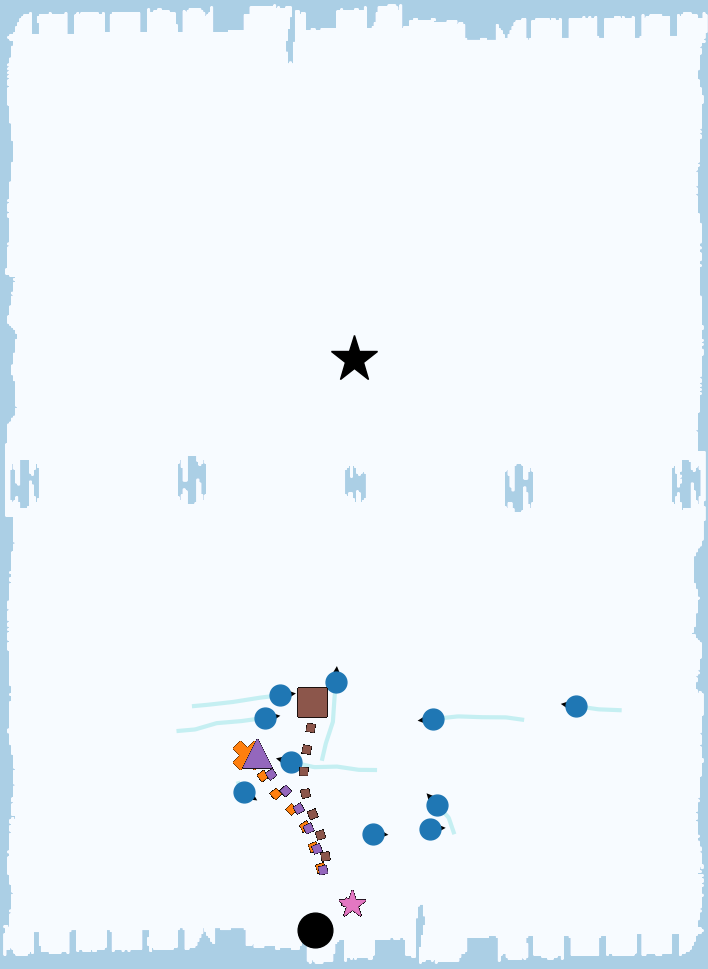} %\caption{$t$ = 5.3s}
    \end{subfigure}
    \begin{subfigure}[t]{0.19\linewidth}
    \includegraphics[width=\linewidth]{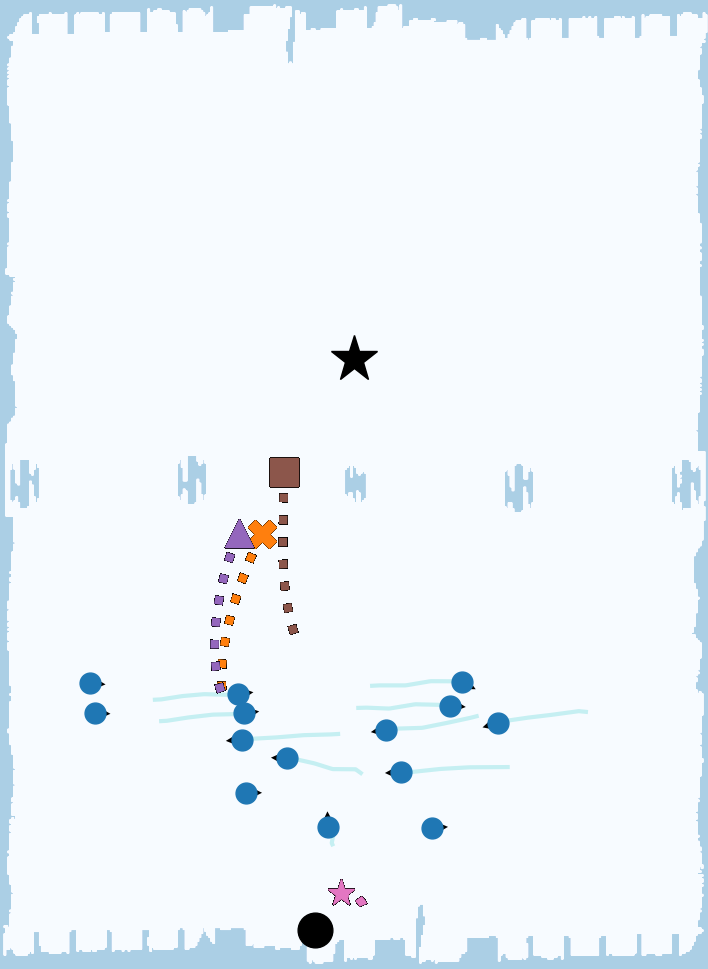} %\caption{$t$ = 10.25s}
    \end{subfigure}
    \begin{subfigure}[t]{0.19\linewidth}
    \includegraphics[width=\linewidth]{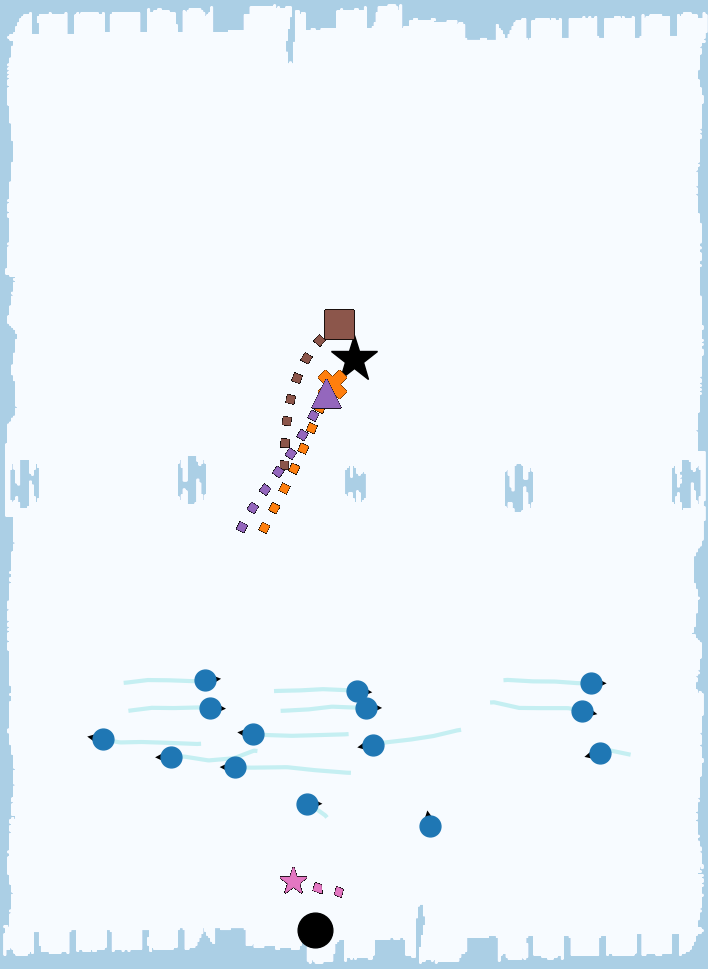} %\caption{$t$ = 13.75s}
    \end{subfigure}
    \begin{subfigure}[t]{0.19\linewidth}
    \includegraphics[width=\linewidth]{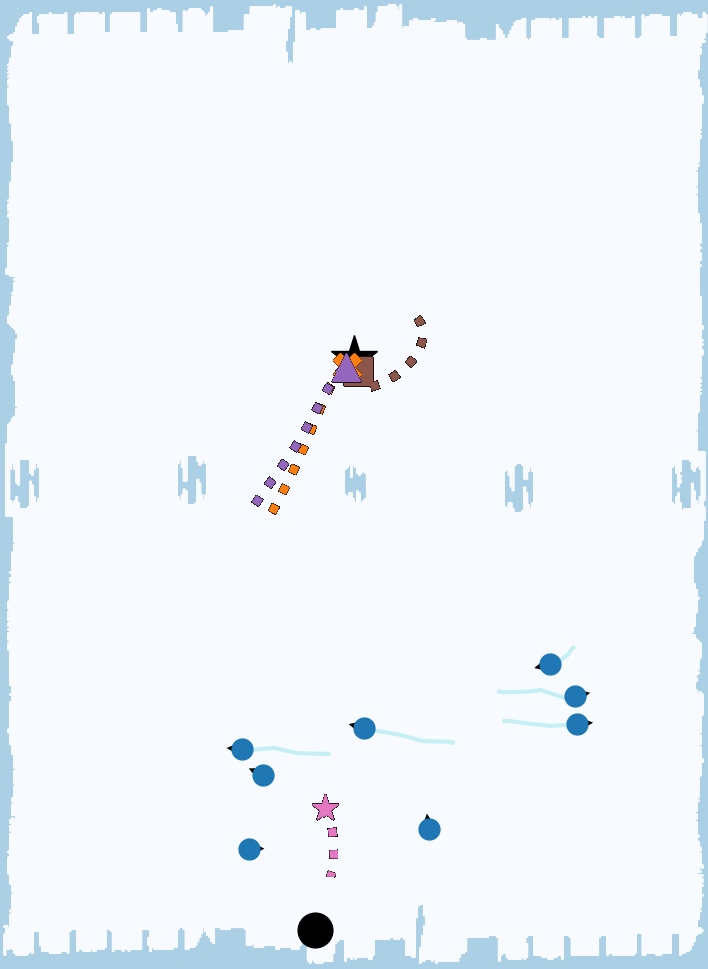} %\caption{$t$ = 25s}
    \end{subfigure}
    \caption{\textbf{Hotel Environment.} Robots cross two lanes of pedestrian traffic. SortCMA and metric tuning perform similarly.}
    %, with only about half of users preferring one over the other. Automatic tuning with CMA and user tuning with SortCMA are both preferable over policies, including user tuning with Bayesian Optimization and learned rewards, as in PrefGP~\cite{Biyik2020}.
    \label{fig:traj_hotel}
\end{figure*}

\begin{figure*}[htb]
    \centering
    %\begin{subfigure}[t]{0.9\linewidth}
    %\includegraphics[width=\linewidth]{figs/traj/legend.png} %\caption{$t$ = 4s}
    %\end{subfigure}
    %\vspace{0.1cm}
    %\begin{subfigure}[t]{0.12\linewidth}
    %\includegraphics[width=\linewidth]{figs/vote_graph_new/Zara_pair.png} 
    %\end{subfigure}
    \begin{subfigure}[t]{0.24\linewidth}
    \includegraphics[width=\linewidth]{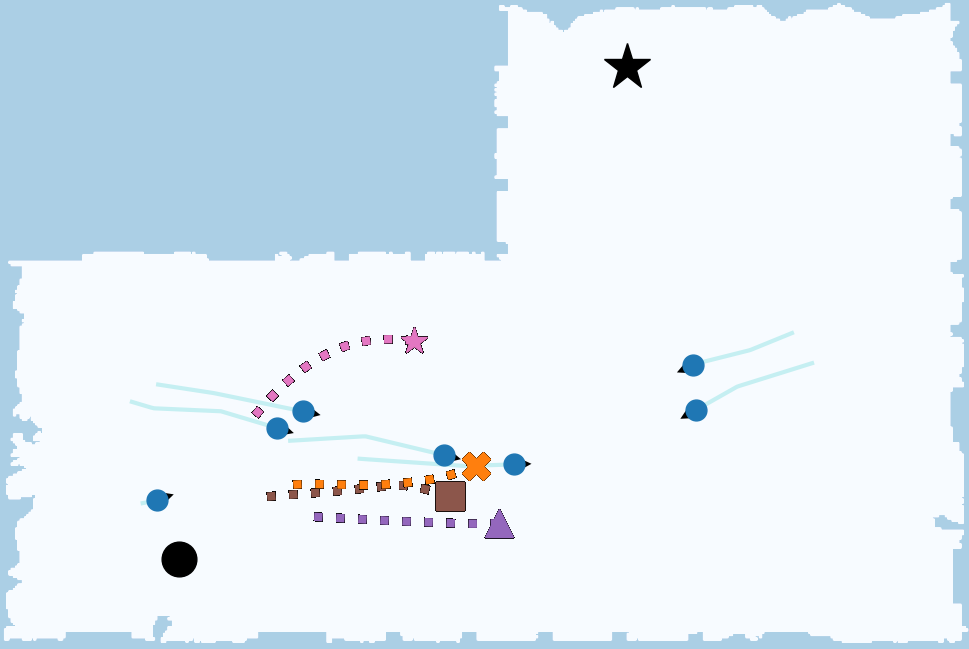} %\caption{$t$ = 6.75s}
    \end{subfigure}
    \begin{subfigure}[t]{0.24\linewidth}
    \includegraphics[width=\linewidth]{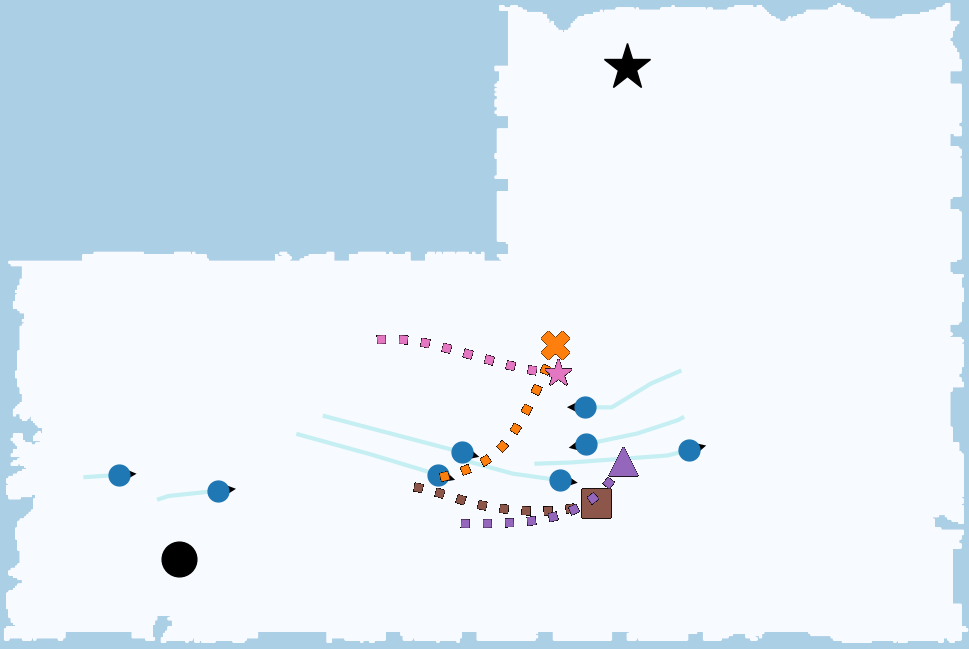} %\caption{$t$ = 9.5s}
    \end{subfigure}
    \begin{subfigure}[t]{0.24\linewidth}
    \includegraphics[width=\linewidth]{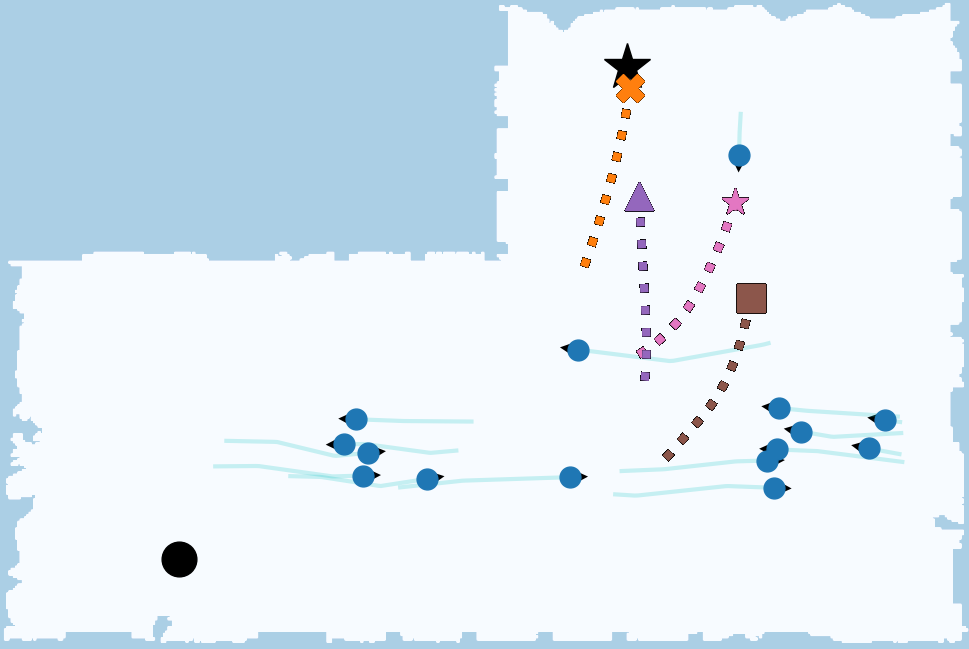} %\caption{$t$ = 14.5s}
    \end{subfigure}
    \begin{subfigure}[t]{0.24\linewidth}
    \includegraphics[width=\linewidth]{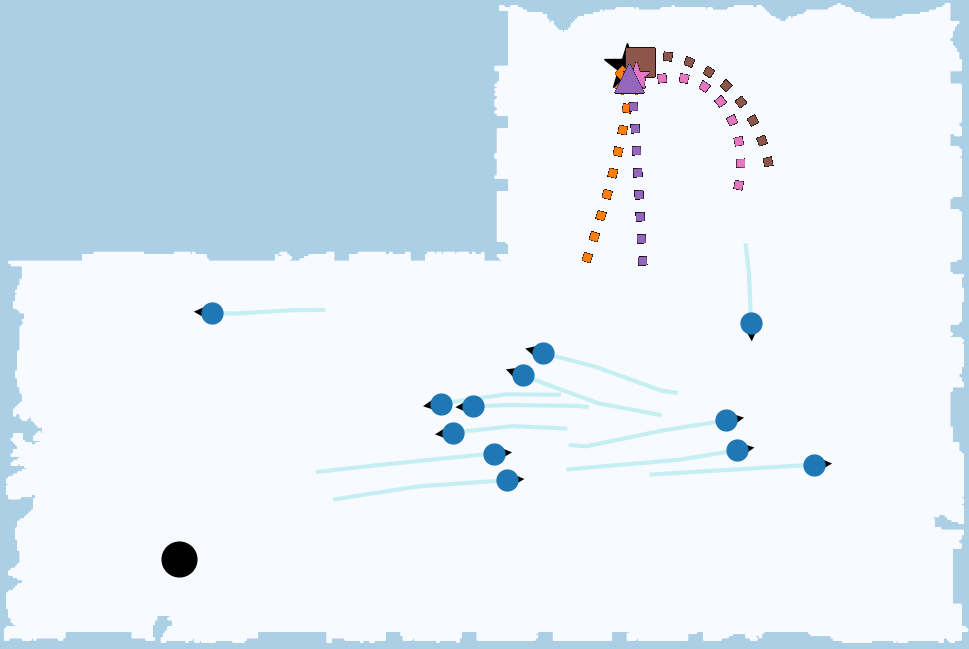} %\caption{$t$ = 20.6s}
    \end{subfigure}
    \caption{\textbf{Zara Environment.} Robots must cut across a traffic at the appropriate time. SortCMA is overall preferred.}
    \label{fig:traj_zara}
\end{figure*}

\begin{figure*}[htb]
    \centering

    %\begin{subfigure}[t]{0.13\linewidth}
    %    \includegraphics[width=\linewidth]{figs/vote_graph_new/Univ_pair.png} 
    %\end{subfigure}
    \begin{subfigure}[t]{0.24\linewidth}
    \includegraphics[width=\linewidth]{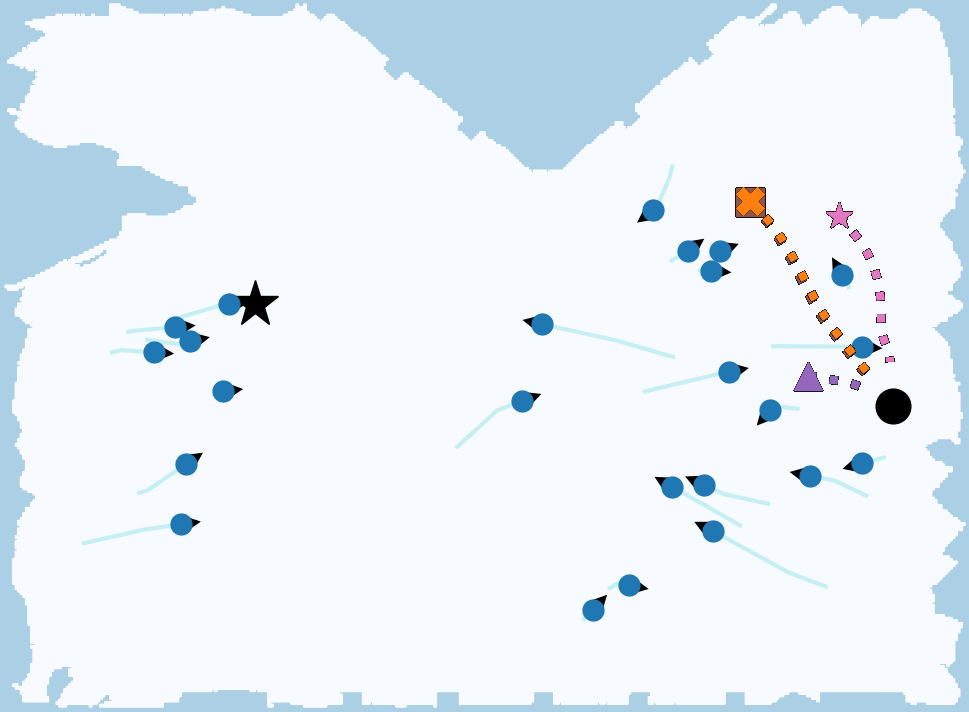} %\caption{$t$ = 5s}
    \end{subfigure}
    \begin{subfigure}[t]{0.24\linewidth}
    \includegraphics[width=\linewidth]{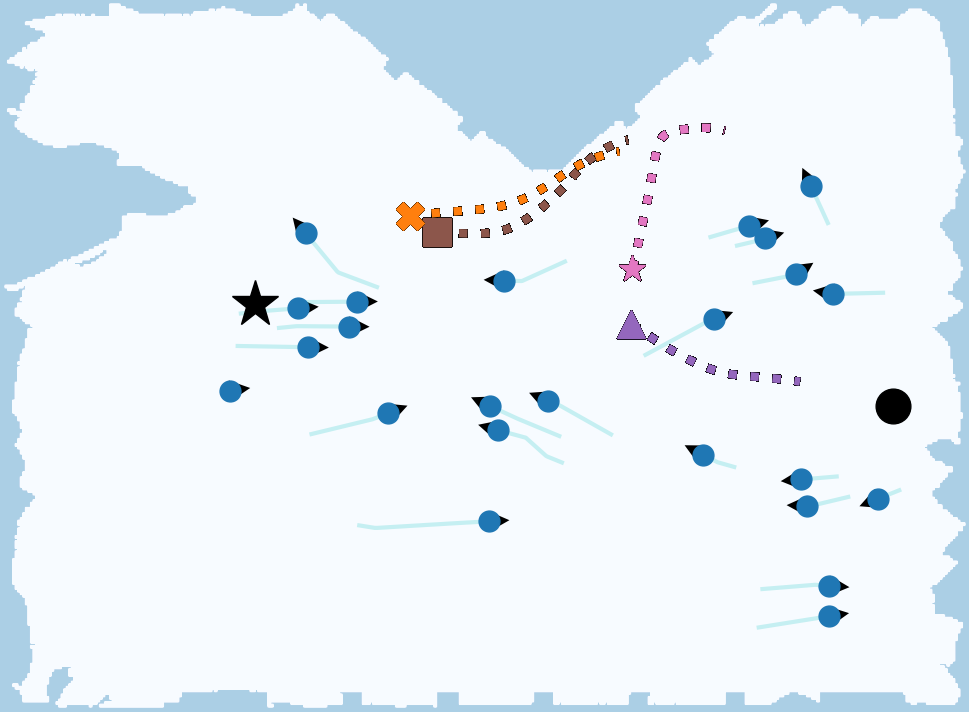} %\caption{$t$ = 11s}
    \end{subfigure}
    \begin{subfigure}[t]{0.24\linewidth}
    \includegraphics[width=\linewidth]{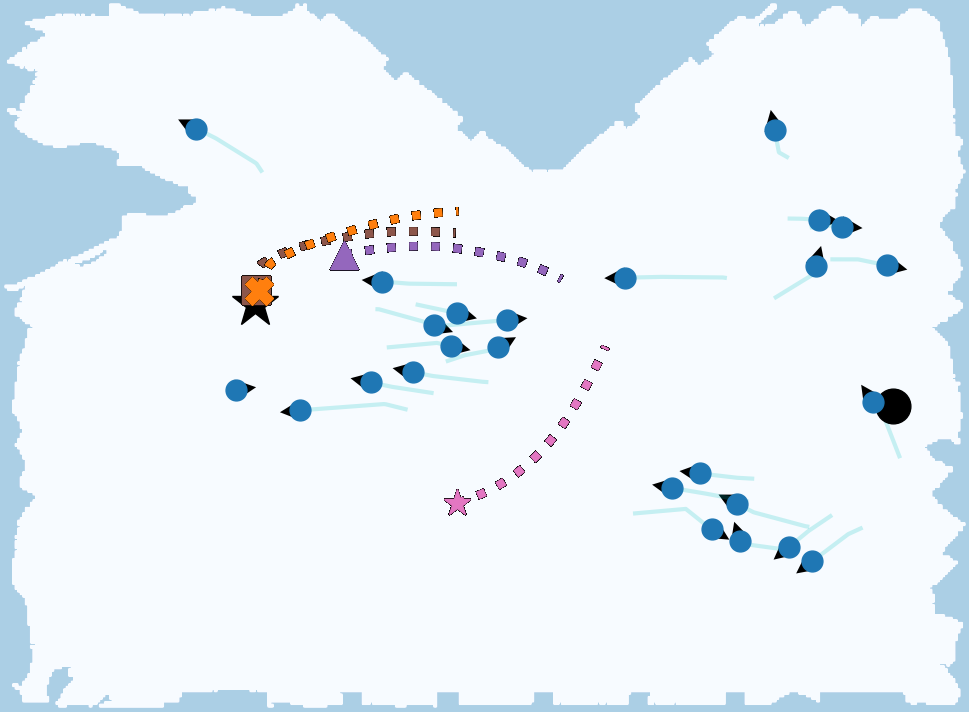} %\caption{$t$ = 16s} 
    \end{subfigure}
    \begin{subfigure}[t]{0.24\linewidth}
    \includegraphics[width=\linewidth]{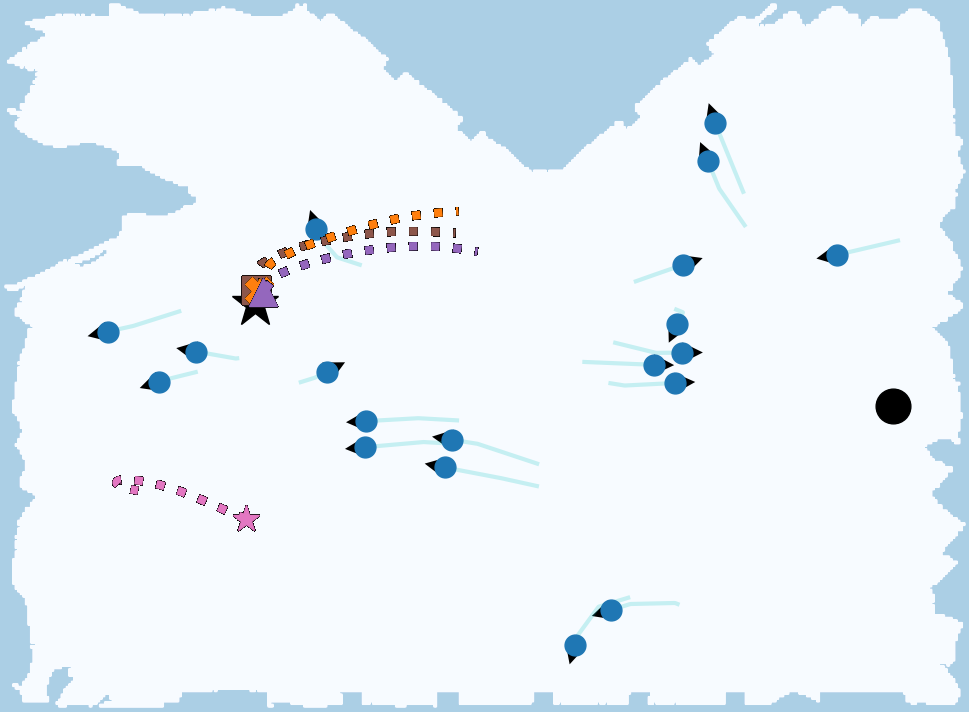} %\caption{$t$ = 25s}
    \end{subfigure} 
    \vspace{0.04cm}

    \centering
    %\begin{subfigure}[t]{0.9\linewidth}
    %\includegraphics[width=\linewidth]{figs/traj/legend.png} %\caption{$t$ = 4s}
    %\end{subfigure}
    %\vspace{0.1cm}
   % \begin{subfigure}[t]{0.19\linewidth}
   % \includegraphics[width=\linewidth]%{figs/vote_graph_new/ETH_pair.png} 
    %\end{subfigure}
    \begin{subfigure}[t]{0.19\linewidth}
    \includegraphics[width=\linewidth]{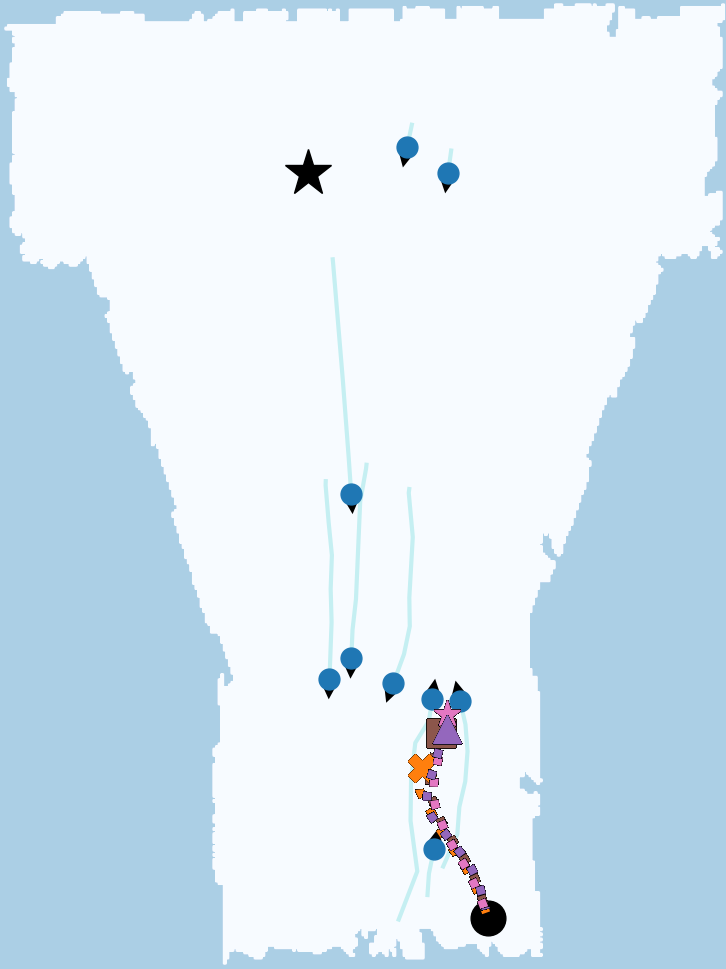} %\caption{$t$ = 4s}
    \end{subfigure}
    \begin{subfigure}[t]{0.19\linewidth}
    \includegraphics[width=\linewidth]{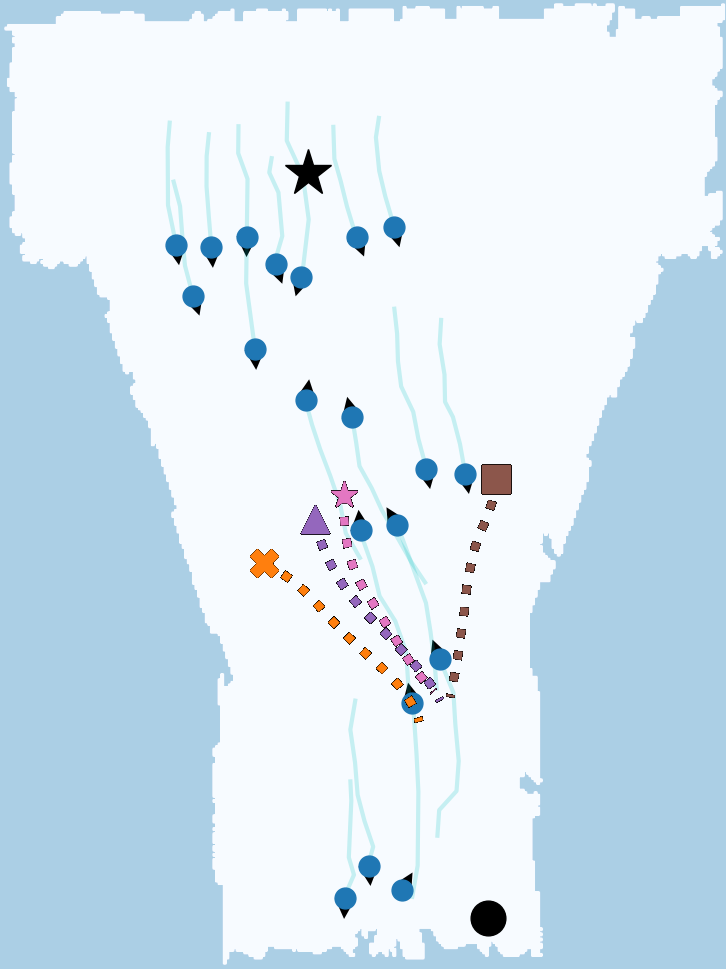} %\caption{$t$ = 8.5s}
    \end{subfigure}
    \begin{subfigure}[t]{0.19\linewidth}
    \includegraphics[width=\linewidth]{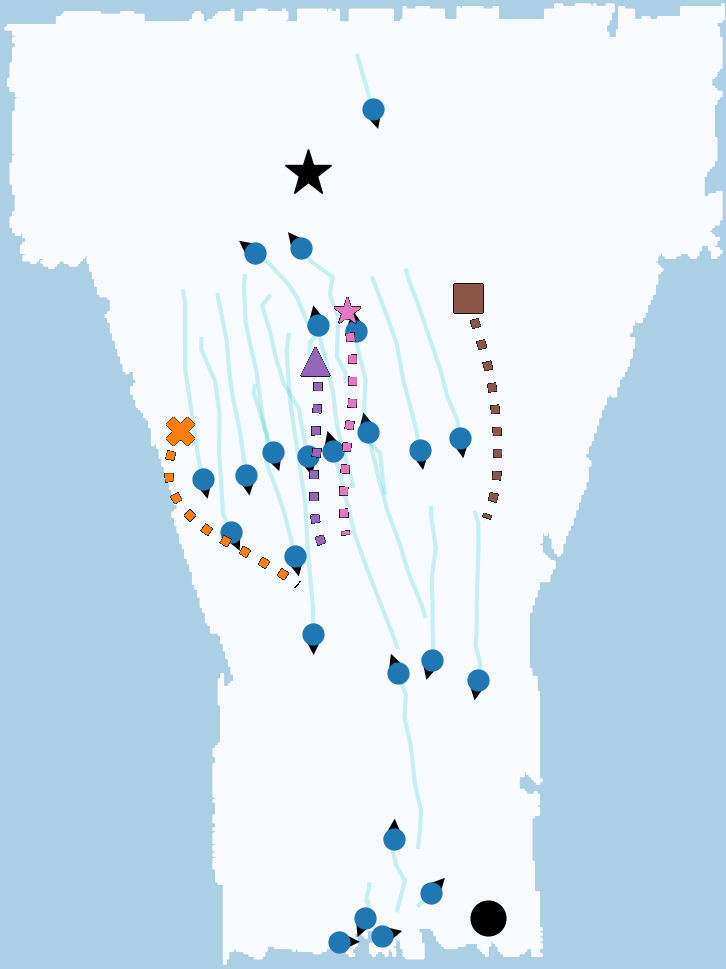} %\caption{$t$ = 11.5s}
    \end{subfigure}
    \begin{subfigure}[t]{0.19\linewidth}
    \includegraphics[width=\linewidth]{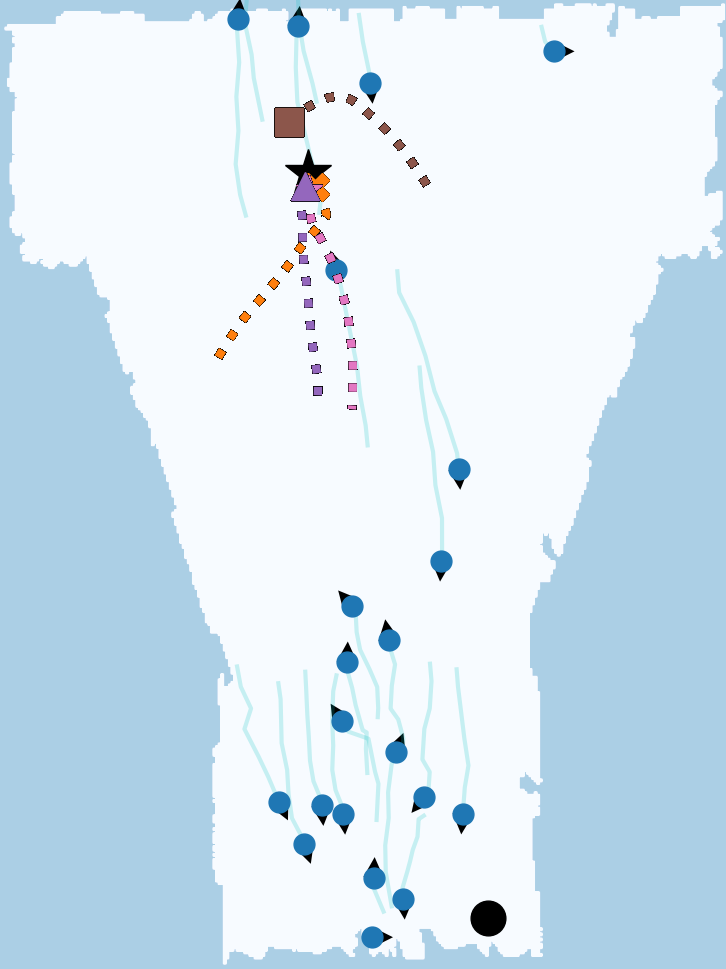} %\caption{$t$ = 17s}
    \end{subfigure}
    \begin{subfigure}[t]{0.19\linewidth}
    \includegraphics[width=\linewidth]{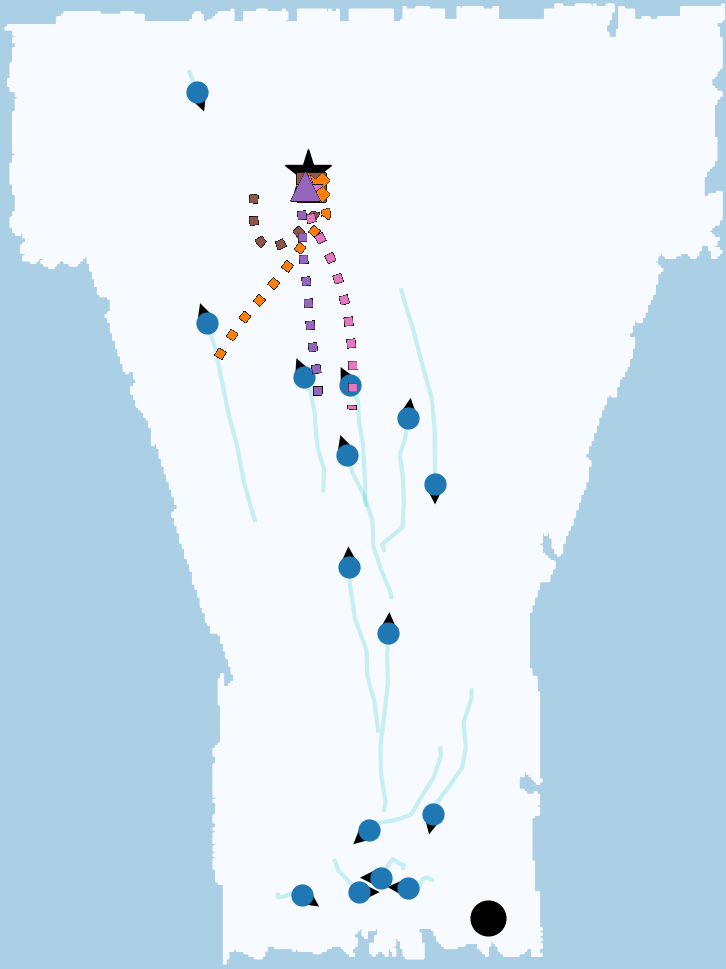} %\caption{$t$ = 21.95s}
    \end{subfigure}

    \caption{\textbf{Univ and ETH Environments.} Robots must cross dense crowds with multiple oncoming pedestrians. }
    \label{fig:traj_univ}
    %We used SortCMA and PrefGP to direct the robot through the middle of the crowd, a policy that most users did not prefer overall.
\end{figure*}

Using SortCMA, we successfully find parameter sets for the classic social forces algorithm ($d = 12$) that achieve our qualitative goals. Visualizations of the resulting trajectories are found in \cref{fig:traj_hotel,fig:traj_univ,fig:traj_zara}.

In the `Hotel' environment (Fig. \ref{fig:traj_hotel}), within seven generations we tuned parameters so that the robot \textit{``wait[s] to cross, and go[es] behind people when possible''}. The SortCMA-tuned robot avoids passing through the pedestrians' forward path, then heads directly towards the goal.

In the `Zara' environment (Fig. \ref{fig:traj_zara}), we show that we can fine-tune for hard-to-enumerate solutions. We tuned the robot to \textit{``follow the crowd, then wait before crossing''} in three generations. SortCMA finds a parameter set that results in the robot crossing the crowd without darting in front of pedestrians, a goal difficult to enumerate with a loss function.

In the `ETH' and `Univ' environments (\cref{fig:traj_univ}), we show that we can achieve specific behaviors with 
preference tuning. Respectively, we tuned parameters so that the robot \textit{``use[s] the inner lane without hitting people''}, and \textit{``navigate[s] safely through the crowd''}. In both cases, after the seventh generation of tuning, the robots use a lane of pedestrian traffic to reach their goal.

%SortCMA is able to generate good trajectories across a variety of environments with relatively few evaluations required. Current social navigation models tend to rely on training based on naturalistic pedestrian data; as geographic differences are significant, deploying across regions requires large amounts of data. Here,  a single developer is able to tune these models quickly. This opens the door to alternate methods of algorithm localization. Instead of large-scale data collection, a few people familiar with local norms could use SortCMA to teach robots how to move in a socially appropriate manner for their region.

\begin{table*}[htbp]
\smallskip
     \caption{\textbf{Pairwise preferences} from ranked order for social navigation with 95\% Clopper-Pearson intervals~\cite{10.1093/biomet/26.4.404}. }
     % , where a $\leq 50\%$ lower bound implies lack of significance. in different environments with Default, Automatic Tuning, and User Tuning (with SortCMA \& PrefGP).
     \begin{subtable}[h]{0.24\textwidth}
        \centering
        \setlength\tabcolsep{1.5pt}

        \begin{tabular}{@{}llll@{} }
\toprule
 Prefer & To & $\mu$ \, [LB, UB] \\ \midrule
\colorbox[HTML]{FFA556}{Metric} & \colorbox[HTML]{B494D0}{CMA} & 55\% [45, 64] \\ 
\colorbox[HTML]{FFA556}{Metric} & \colorbox[HTML]{B68378}{Def} & 76\% [67, 84] \\ 
\colorbox[HTML]{FFA556}{Metric} & \colorbox[HTML]{EB9FD4}{GP} & 88\% [80, 93] \\ 
\colorbox[HTML]{B494D0}{CMA} & \colorbox[HTML]{B68378}{Def} & 77\% [68, 84] \\ 
\colorbox[HTML]{B494D0}{CMA} & \colorbox[HTML]{EB9FD4}{GP} & 87\% [79, 92] \\ 
\colorbox[HTML]{B68378}{Def} & \colorbox[HTML]{EB9FD4}{GP} & 73\% [64, 81] \\ \bottomrule
\end{tabular}
        \caption{Hotel}
       \label{tab:week1}
    \end{subtable}
     \begin{subtable}[h]{0.24\textwidth}
        \centering
        \setlength\tabcolsep{1.5pt}
\begin{tabular}{@{}llll@{} }
\toprule
 Prefer & To & $\mu$ \, [LB, UB] \\ \midrule
\colorbox[HTML]{B494D0}{CMA} & \colorbox[HTML]{FFA556}{Metric} & 68\% [59, 77] \\ 
\colorbox[HTML]{B494D0}{CMA} & \colorbox[HTML]{B68378}{Def} & 81\% [73, 88] \\ 
\colorbox[HTML]{B494D0}{CMA} & \colorbox[HTML]{EB9FD4}{GP} & 91\% [84, 96] \\ 
\colorbox[HTML]{FFA556}{Metric} & \colorbox[HTML]{B68378}{Def} & 53\% [43, 63] \\ 
\colorbox[HTML]{FFA556}{Metric} & \colorbox[HTML]{EB9FD4}{GP} & 82\% [74, 89] \\ 
\colorbox[HTML]{B68378}{Def} & \colorbox[HTML]{EB9FD4}{GP} & 74\% [65, 82] \\ \bottomrule
\end{tabular}
        \caption{Zara}
        \label{tab:week2}
     \end{subtable}
         \begin{subtable}[h]{0.24\textwidth}
        \centering
        \setlength\tabcolsep{1.5pt}

\begin{tabular}{@{}llll@{} }
\toprule
 Prefer & To & $\mu$ \, [LB, UB] \\ \midrule
\colorbox[HTML]{FFA556}{Metric} & \colorbox[HTML]{B68378}{Def} & 76\% [67, 84] \\ 
\colorbox[HTML]{FFA556}{Metric} & \colorbox[HTML]{B494D0}{CMA} & 73\% [64, 81] \\ 
\colorbox[HTML]{FFA556}{Metric} & \colorbox[HTML]{EB9FD4}{GP} & 68\% [59, 77] \\ 
\colorbox[HTML]{B68378}{Def} & \colorbox[HTML]{B494D0}{CMA} & 69\% [60, 77] \\ 
\colorbox[HTML]{B68378}{Def} & \colorbox[HTML]{EB9FD4}{GP} & 66\% [57, 75] \\ 
\colorbox[HTML]{B494D0}{CMA} & \colorbox[HTML]{EB9FD4}{GP} & 55\% [45, 64] \\ \bottomrule
\end{tabular}
        \caption{Univ}

        \label{tab:week3}
     \end{subtable}
         \begin{subtable}[h]{0.24\textwidth}
        \centering
        \setlength\tabcolsep{1.5pt}

                \begin{tabular}{@{}llll@{} }
\toprule
 Prefer & To & $\mu$ \, [LB, UB] \\ \midrule
\colorbox[HTML]{B68378}{Def} & \colorbox[HTML]{FFA556}{Metric} & 56\% [46, 65] \\ 
\colorbox[HTML]{B68378}{Def} & \colorbox[HTML]{B494D0}{CMA} & 81\% [72, 87] \\ 
\colorbox[HTML]{B68378}{Def} & \colorbox[HTML]{EB9FD4}{GP} & 88\% [80, 93] \\ 
\colorbox[HTML]{FFA556}{Metric} & \colorbox[HTML]{B494D0}{CMA} & 73\% [64, 81] \\ 
\colorbox[HTML]{FFA556}{Metric} & \colorbox[HTML]{EB9FD4}{GP} & 81\% [73, 88] \\ 
\colorbox[HTML]{B494D0}{CMA} & \colorbox[HTML]{EB9FD4}{GP} & 64\% [54, 73] \\ \bottomrule
\end{tabular}
        \caption{ETH}

        \label{tab:week4}
     \end{subtable}

     \label{tab:pairwise}
\end{table*}

\subsection{User Study}\label{sec:userstudy}
To compare the learned parameter sets, we design an internet survey-based user study. The study and recruitment were approved by our Institutional Review Board.

We use four methods to generate parameter sets, again for the social forces algorithm. They are as follows:
\begin{itemize}
    \item \textbf{\textit{Default:}} Parameter settings as given in SocNavBench that achieved the most success in the original benchmark set. These were originally chosen based on \cite{Moussad2009}, a study of pedestrian behavior in crowds, and are intended to produce human-like robot behavior.
    \item \textbf{\textit{Metric:}} Parameter settings found by optimizing over a hand-designed loss function. We use the geometric mean of the following metrics in SocNavBench: a) success (if goal was reached); b) goal traversal ratio (partial success); c) path irregularity (measuring efficiency); d) travel time; e) thresholded distance to pedestrians; and f) thresholded time to collision. Other than the design of the loss function, this does not require user feedback.
    \item \textbf{\textit{PrefGP:}} Parameter settings tuned by user preference using the active querying algorithm from \cite{Biyik2020}. We use pairwise preferences to fit a regularized logistic regression to SocNavBench metrics, and pick the best configuration based on the resulting reward function.
    \item \textbf{\textit{SortCMA:}} Parameter settings tuned using our method. 
\end{itemize}

Models are tuned for each environment individually. The optimizers ran for a total of 165 sampled trajectories ($\approx14N$). We pre-specify a qualitative goal for each environment before tuning the preference-based parameter sets, and choose the better of each trajectory pair based on subjective similarity to that goal. When two trajectories appear the same, we use a heuristic which rates them based on the same loss function from the Metric parameter set.

% HOTEL: Our preference-based SortCMA and metric parameter sets both avoid passing through the pedestrians' forward path. The default parameter set guides the robot directly through the crowd. PrefGP is unable to reach the goal within a reasonable time. 

% ZARA: Preference-tuned, default, and metric-tuned all use the same strategy to reach the goal, but cross the crowd at different points.

In the first part of the study, participants view single robots navigating through each scenario (16 videos, one for each optimized parameter set and environment pair). They then rate trajectories with 7-point Likert-style responses (strongly disagree to strongly agree) to the following statements: \textit{I found the robot to be competent} and \textit{I am wary of the robot}. In the second part of the study, we show all four trajectories for a given environment in a single video, and ask users to \textit{rank the robots based on how [they] think the robots should move}. Robots cannot be ranked equally. For ease of viewing, collisions with pedestrians are not marked in this phase.

We collected results from 150 Prolific workers over the age of 18 based in the United States. Of the 150 participants, we exclude 5 who failed our attention check and 5 who took over an hour to finish the survey (which had a median completion time of 15 minutes). Because we are interested in evaluations of robot behaviors over the entire trajectory, we split the remaining participants into two groups: those who did not finish watching each video and those who only occasionally skipped the ending of a video. We ran a Mann-Whitney-Wilcoxon test over the Likert responses of both groups, and found differences in responses ($p < 0.001$). We focus our analysis on the set of completionists, with $n = 113$.

Survey responses for competence and safety were analyzed with a two-way repeated-measures ANOVA across optimizer and environment. Results for both competence and wariness showed significant results for optimizer, environment, and their interaction (\textit{p} < 0.0001 in all cases). Competence and safety were somewhat correlated (\textit{r}(111) = 0.646, \textit{p} < 0.001): optimizers viewed as competent were also viewed as safe.

\subsection{User Study Results}\label{sec:userstudyresults}
Our user study includes two evaluations: independent Likert scale ratings of each trajectory, shown with person-robot collisions; and ranked preferences from a video where all robot paths are shown, but collisions are not. %Quantitative results of the Likert-style ratings are shown in \cref{fig:vote_graphs}, and the ranked preferences in \cref{tab:pairwise}. 

Under the Likert-style evaluation (\cref{fig:vote_graphs}), where collisions were visualized, our SortCMA tuned configuration was always among the most preferred. It was either alone (`Hotel', `Univ') or within one standard error of another method (`Zara', `ETH'). Our other preferential tuning baseline (PrefGP) was always among the least preferred, either alone (`Hotel', `Univ', `ETH') or tied with others (`Zara'). In `Hotel', SortCMA's configurations were highly preferred to the other three. In `Zara', all methods performed similarly. In `Univ', SortCMA slightly outperformed the Default and Metric baselines, and PrefGP performed poorly. In `ETH', SortCMA and Metric outperformed the Default while PrefGP  performed poorly. Our study had separate Wariness and Competence questions, and we found that both questions captured similar performance preferences among trajectories.

Under the ranked order evaluation (\cref{tab:pairwise}), where collisions were not visualized, our results are less consistent across environments. In `Zara', where Likert scores were all similar, SortCMA is significantly preferred to other methods, highlighting that rank order analysis can extract finer differences \cite{10.2307/2749259,10.2307/3150506}. In `Hotel', the lack of collision visualization led to a statistical tie between SortCMA and Metric as winners, despite a clear preference towards the former in Likert scores. In both `Univ' and `ETH', SortCMA tuned trajectories were not the most preferred under ranked choice evaluation, which lacked visualization of collisions. %Metric was preferred in the former and Defaults in the latter. 

While our tuned preferences for social navigation do not match those of the survey population, our user study  demonstrated that preference tuning with SortCMA can readily generate social forces models that produce robot trajectories that are broadly preferred in a survey population. 

\section{Discussion}\label{sec:discuss}

\subsection{Linear Reward Learning}\label{sec:learnedreward}
\begin{figure}[htbp]
\centering
\includegraphics[width=\linewidth]{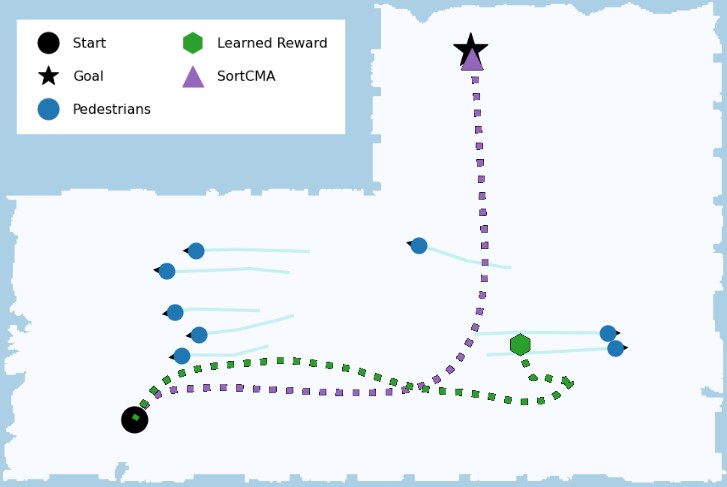}
\caption{\textbf{Learned Rewards} can be insufficient (\cref{sec:learnedreward})}
\label{fig:learned_reward}
\end{figure}

Based on the preferences shown during the SortCMA optimization, we fit a logistic regression-based classifier of predicted user feedback using all available metrics from SocNavBench as features. We optimize for a new set of parameters over the resulting reward function. However, these learned reward trajectories are unable to reproduce the desired behavior. For example, in Fig. \ref{fig:learned_reward}, the parameter set tuned via learned reward does not make it to the goal. This behavior was similar across environments. %While in some cases good linear rewards can be learned~\cite{Biyik2022,Biyik2020}, in our application, even with the many metrics of SocNavBench, linear models of reward were insufficient.

\subsection{Viability of Batch Learning}\label{sec:batch}
We would like to use preferences to tune a generalized set of parameters across different scenarios; however, simultaneous pairwise comparison across multiple environments quickly becomes difficult for users to evaluate. As a surrogate, we compare optimization results for metric tuning under different evaluation schedules.

\begin{figure}[htbp]
\smallskip
\smallskip

\centering
\includegraphics[width=\linewidth]{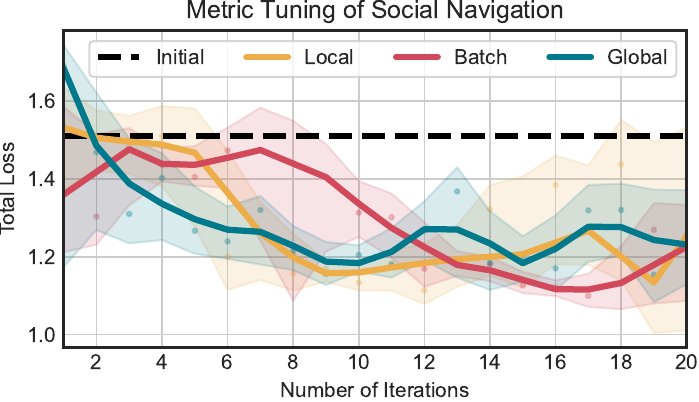}
\caption{\textbf{Schedule Robustness} appears to exist in SortCMA.}
\label{fig:glob_loc_comp}
\end{figure}

We use three schedules during these ablations: global, which optimizes based on the total cost across all environments; local, which is the sum cost of the four individually tuned parameter sets; and batch, which shuffles the four environments then cycles through the set. In all cases, the cost decreases with iterations, and the cost is similar across schedules (Fig. \ref{fig:glob_loc_comp}). This implies that the batch method is a viable way of allowing a user to  tune a global optimizer.

In this setting, recent techniques from federated optimization~\cite{FL_Bandit_Neurips20,FL_ES_paper21,FL_ES_paper22} may potentially be adapted to our setting, with environments as a separate instances or objectives. 

\section{Conclusion}\label{sec:conc}
We demonstrate the SortCMA algorithm can optimize algorithm configurations in many robotics areas. The algorithm is particularly effective where metrics are unavailable or difficult to obtain and formalize. As a secondary result, we show how social navigation algorithms can directly be tuned using SortCMA to produce acceptable behaviors.


\bibliographystyle{IEEEtran}
\bibliography{IEEEabrv,mybibfile}

\end{document}